\documentclass[sigconf]{acmart}

\usepackage{latexsym}
\usepackage{multirow} 
\usepackage{booktabs}
\usepackage{makecell}
\usepackage{rotating}
\usepackage{comment}
\usepackage{enumitem}

\newcommand{\green}[1]{{\color[HTML]{2ca02c}\textbf{#1}}}

\newcommand{\xiaolan}[1]{\green{[xiaolan] #1}}

\newcommand{\todo}[1]{{\color{red} {[ToDo] #1}}}
\newcommand{\smallurl}[1]{{\small{\url{#1}}}}

\AtBeginDocument{%
  \providecommand\BibTeX{{%
    \normalfont B\kern-0.5em{\scshape i\kern-0.25em b}\kern-0.8em\TeX}}}

\copyrightyear{2022}
\acmYear{2022}
\setcopyright{acmcopyright}
\acmConference[SIGIR '22]{Proceedings of the 45th International ACM SIGIR Conference on Research and Development in Information Retrieval}{July 11--15, 2022}{Madrid, Spain}
\acmBooktitle{Proceedings of the 45th International ACM SIGIR Conference on Research and Development in Information Retrieval (SIGIR '22), July 11--15, 2022, Madrid, Spain}
\acmPrice{15.00}
\acmDOI{10.1145/3477495.3532676}
\acmISBN{978-1-4503-8732-3/22/07}

\begin{document}
\fancyhead{}

\title{Beyond Opinion Mining:\\Summarizing Opinions of 
Customer Reviews}

\author{Reinald Kim Amplayo}
\affiliation{
\institution{Google Research}
\country{}
}
\email{reinald@google.com}

\author{Arthur Bra\v{z}inskas}
\affiliation{
\institution{Google Research}
\country{}
}
\email{abrazinskas@google.com}

\author{Yoshi Suhara}
\affiliation{
\institution{Grammarly}
\country{}
}
\email{yoshi.suhara@grammarly.com}

\author{Xiaolan Wang}
\affiliation{
\institution{Megagon Labs}
\country{}
}
\email{xiaolan@megagon.ai}

\author{Bing Liu}
\affiliation{
\institution{University of Illinois at Chicago}
\country{}
}
\email{liub@uic.edu}

\renewcommand{\shortauthors}{Amplayo et al.}

\begin{abstract}
Customer reviews are vital for making purchasing decisions in the Information Age. Such reviews can be automatically summarized to provide the user with an overview of opinions. In this tutorial, we present various aspects of opinion summarization that are useful for researchers and practitioners. First, we will introduce the task and major challenges. Then, we will present existing opinion summarization solutions, both pre-neural and neural. We will discuss how summarizers can be trained in the unsupervised, few-shot, and supervised regimes. Each regime has roots in different machine learning methods, such as auto-encoding, controllable text generation, and variational inference. Finally, we will discuss resources and evaluation methods and conclude with the future directions. This three-hour tutorial will provide a comprehensive overview over major advances in opinion summarization. The listeners will be well-equipped with the knowledge that is both useful for research and practical applications.\footnote{All the materials are available at: 
\\ \url{https://github.com/abrazinskas/sigir2022-opinion-summarization-tutorial}.}

\end{abstract}

\ccsdesc[500]{Information systems~Summarization}
\ccsdesc[300]{Information systems~Sentiment analysis}
\ccsdesc[100]{Information systems~Information extraction}
\keywords{opinion mining, opinion summarization}

\maketitle

\section{Introduction}

People in the Information Age read reviews from online review websites
when making decisions to buy a product or use a service. The proliferation of such reviews has driven research on opinion mining \cite{hu2006opinion,pang2008opinion}, where the ultimate goal is to glean information from multiple reviews so that users can make decisions more effectively. Opinion mining has assumed several facets in its history: among others, there are sentiment analysis \cite{pang2002thumbs}, that reduces a single review into a sentiment label, 
opinion extraction \cite{mukherjee-liu-2012-aspect}, that produces a list of aspect-sentiment pairs representing opinions mentioned in the reviews, 
and most notably
\textit{opinion summarization} \cite{wang-ling-2016-neural}, which creates a textual summary of opinions that are found in multiple reviews about a certain product or service.
Opinion summarization is arguably the most effective solution for opinion mining, especially when assisting the user in making decisions. Specifically, textual opinion summaries provide users with information that is both more concise and more comprehensible compared to other alternatives. Thus, opinion mining research on the IR community
has geared its focus towards opinion summarization in recent years (see Table~\ref{tab:sols}).

The task of summarizing opinions in multiple reviews can be divided into two subtasks: opinion retrieval and summary generation. Opinion retrieval selects opinions from the reviews that are salient and thus need to be included in the summary.
Summary generation produces a textual summary given the retrieved opinions that is concise yet informative and comprehensible for users to read and make decisions effectively.
The summary can be generated from scratch with possibly novel tokens (i.e., \textit{abstractive} summarization; \cite{opinosis,meansum}) or spans of text directly extracted from the input (i.e., \textit{extractive} summarization; \cite{hu2004mining2,mate_mt}).
Traditionally, these subtasks correspond to a pipeline of natural language generation models \cite{mckeown1992text,carenini-etal-2006-multi,wang-ling-2016-neural} where opinion retrieval and summary generation are treated as content selection and surface realization tasks, respectively.
Thanks to advancements in neural networks, most of the recent methods use an end-to-end approach \cite{meansum,copycat,fewsum} where both opinion retrieval and summary generation are done by a single model optimized to produce well-formed and informative summaries.

There are two broad types of challenges in opinion summarization: 
\textit{annotated data scarcity} and \textit{usability}. As reviews-summary pairs are expensive to create, this has resulted in annotated dataset scarcity. However, the exceptional performance of neural networks for text summarization is mostly driven by large-scale supervised training~\citep{rush2015neural,zhang2020pegasus}, which makes opinion summarization challenging. The second challenge -- usability -- stems from a number of practical requirements for industrial applications. First, for real-world products and service we often need to summarize many thousands of reviews. This is largely infeasible due to the high computational and memory costs of modelling that many reviews with neural architectures~\citep{Beltagy2020Longformer}. Second, state-of-the-art text summarizers are prone to hallucinations~\citep{maynez-etal-2020-faithfulness}. In other words, a summarizer might mistakenly generate a summary with information not covered by input reviews, thus misinforming the user. Third, generic summaries often cannot address specific user needs. This, in turn, calls for ways to learn summarizers producing personalized summaries. 

This opens exciting avenues to develop methods for solving these major challenges in opinion summarization. In this light, the aim of the tutorial is to inform interested researchers and practitioners, especially in opinion mining and text summarization, about recent and ongoing efforts to improve the state of the art and make opinion summarization systems useful in real-world scenarios. And the tutorial will make the audience well-equipped for addressing these challenges in terms of methods, ideas, and related work.

\section{Tutorial Content and Outline}

The tutorial will be 3 hours long and consist of the following 
five parts, which we describe in detail below.

\subsection{Part I: Introduction [30 min]}
Opinion summarization~\cite{hu2006opinion, titov2008joint, kim2011comprehensive} focuses on summarizing opinionated text, such as customer reviews, and has been actively studied by researchers from the natural language processing and data mining community for decades. There are two major types of opinion summaries: non-textual summaries, such as aggregated ratings~\cite{lu2009rated}, aspect-sentiment tables~\cite{titov2008joint}, and opinion clusters~\cite{hu2004mining2}, and textual summaries, which often consist of a short text. Compared to non-textual summaries, which may confuse users due to their complex formats, textual summaries are considered much more  user-friendly~\cite{murray2017opinion}. Thus, in recent years, the considerable research interest in opinion summarization has shifted towards textual opinion summaries. In this tutorial, we will also focus on recent solutions for generating textual opinion summaries.

Like single document summary \cite{rush2015neural,see2017get}, textual opinion summary can also be either extractive or abstractive. However, unlike single document summarization, opinion summarization can rarely rely on gold-standard summaries at training time due to the lack of large-scale training examples in the form of review-summary pairs. Meanwhile, the prohibitively many and redundant input reviews also pose new challenges for the task. 

In this part of the tutorial, we will first describe the opinion summarization task, its history, and the major challenges that come with the task. We will then provide a brief overview of existing opinion summarization solutions.

\begin{table}[]
    \small
    \centering
    \caption{Opinion summarization solutions that will be covered in this tutorial. A dagger $\dagger$ denotes that the solution also leverages weak supervision.}
    \vspace{-2mm}
    \begin{tabular}{@{~}p{8.3cm}@{~}}
        \hline
        \multicolumn{1}{c}{\textbf{Pre-Neural Solutions}} \\
        \hline
        \textbf{Extractive}:\\LexRank~\cite{lexrank}, TextRank~\cite{textrank}, 
         MEAD~\cite{carenini-etal-2006-multi},
         Wang et.al~\cite{wang-etal-2014-query} \\
         \textbf{Abstractive}:\\Opinosis~\cite{opinosis},
         SEA~\cite{carenini-etal-2006-multi},
         Gerani et.al~\cite{gerani2014abstractive} \\
         \hline
         \multicolumn{1}{c}{\textbf{Autoencoders}} \\
         \hline
         \textbf{Extractive}: \\MATE+MT$^\dagger$~\cite{mate_mt},
        Mukherjee et.al$^\dagger$~\cite{mukherjee2020read},
        ASPMEM$^\dagger$~\cite{aspmem},
        QT~\cite{qt} \\
        \textbf{Abstractive}:\\MeanSum~\cite{meansum},
        Coavoux et.al~\cite{coavoux2019unsupervised},
        OpinionDigest$^\dagger$~\cite{opiniondigest},
        RecurSum~\cite{recursum},
        MultimodalSum~\cite{multimodalsum},
        COOP~\cite{coop} \\
        \hline
        \multicolumn{1}{c}{\textbf{Synthetic Training}} \\
        \hline
        \textbf{Abstractive}:\\Copycat~\cite{copycat},
        DenoiseSum~\cite{denoisesum},
        MMDS~\cite{mmds},
        Elsahar et.al$^\dagger$~\cite{elsahar2021self},
        Jiang et.al~\cite{Jiang2021},
        PlanSum~\cite{plansum},
        TransSum~\cite{transsum},
        AceSum~\cite{acesum}, ConsistSum~\cite{ke2022consistsum},
        LSARS~\cite{pan2020large}\\
        \hline
        \multicolumn{1}{c}{\textbf{Low-Resource}} \\
        \hline
        \textbf{Abstractive}: \\
        Wang et.al~\cite{wang-ling-2016-neural},
        FewSum~\cite{fewsum},
        AdaSum~\cite{brazinskas2022efficient},
        PASS~\cite{pass},
        SelSum~\cite{selsum},
        CondaSum~\cite{condasum},
        Wei et.al~\cite{Wei2021}\\
        \hline
    \end{tabular}
    \label{tab:sols}
    \vspace{-0.2cm}
\end{table}

\subsection{Part II: Solutions To Data Scarcity [90 min]}

In this part of the tutorial, we will present multiple existing opinion summarization models, as also summarized in Table~\ref{tab:sols}. These models attempt to solve the annotated data scarcity problem and are classified into four parts: pre-neural models, autoencoder-based models, models that use synthetic data, and models that leverage low-resource annotated data.

\subsubsection{Autoencoders [30/90 min]}\label{subsubsec:autoencoder}
Due to the lack of training examples, one major approach is to use autoencoders for unsupervised opinion summarization. The autoencoder model consists of an encoder that transforms the input into latent representations and a decoder that attempts to \textit{reconstruct} the original input using a reconstruction objective. It has a wide range of applications in both 
CV and NLP communities~\cite{hinton2011transforming,Kingma2014AutoEncodingVB,bowman-etal-2016-generating}. Autoencoders can also help models obtain better text representations, which allows easier text clustering, aggregation, and selection. Thus, it benefits both extractive and abstractive solutions. In this tutorial, we will first introduce the basics of autoencoders and then describe how to use autoencoders for both extractive and abstractive opinion summarization.

\subsubsection{Synthetic Dataset Creation [30/90 min]}
\label{subsubsec:synthetic}

{The supervised training of high-capacity models on large datasets containing hundreds of thousands of document-summary pairs is critical to the recent success of deep learning techniques for abstractive summarization \citep{rush2015neural,see2017get}}.
The absence of human-written summaries in a large-scale calls for creative ways to synthesize datasets for {supervised training of abstractive summarization models}. Customer reviews, available in large quantities, can be used to create synthetic datasets for training. Such datasets are created by sampling one review as a pseudo-summary, and then selecting or generating a subset of reviews as input to be paired with the pseudo-summary. Subsequently, the summarizer is trained in a supervised manner to predict the pseudo-summary given the input reviews. This \textit{self-supervised} approach, as has been shown in a number of works [\citealp{zhang2020pegasus}, \textit{inter alia}], is effective for training summarizers to generate abstractive opinion summamaaries. In this tutorial, we will introduce various techniques to create synthetic datasets, contrast them, and present results achieved by different works.

\subsubsection{Low-Resource Learning [30/90 min]}\label{subsubsec:fewshot}
Modern deep learning methods rely on large amounts of annotated data for training. Unlike synthetic datasets, automatically created from customer reviews, annotated datasets require expensive human effort. Consequently, only datasets with a handful of human-written summaries are available, which lead to a number of few-shot models. These models alleviate annotated data-scarcity using specialized mechanisms, such as parameter subset fine-tuning and summary candidate ranking. An alternative to human-written are editor-written summaries that are scraped from the web and linked to customer reviews. This setup is challenging because each summary can have hundreds of associated reviews. In this tutorial, we will present both methods that are few-shot learners and that scale to hundreds of input reviews. 

\vspace{-0.5mm}
\subsection{Part III: Improving Usability [30 min]}\label{subsubsec:advanced}

In order to make opinion summarizers more useful in industrial settings, a number of features need to be improved. In this part of the tutorial, we will discuss the following three major features and recent solutions the community has proposed:

\begin{itemize}[leftmargin=*]
    \setlength\itemsep{1mm}
    \item \textbf{Scalability: }
    The ability to handle a massive number of input reviews. To handle large scale input, the ability to retrieve salient information, e.g., reviews or opinions, becomes a important yet challenging feature for opinion summarization solutions. 
    \item \textbf{Input Faithfulness: }
    The ability of a summarizer to generate summaries covered in content by input reviews. In other words, the summarizer should not confuse entities or introduce novel content into summaries.
    
    \item \textbf{Controllability: }
    The ability to produce constrained summaries, such as a hotel summary that only includes room cleanliness  or a product summary that only covers the negative opinions. 
\end{itemize}

\subsection{Part IV: Evaluation and Resources [20 min]}

As is common in other areas of natural language processing, in opinion summarization, researchers often rely on automatic metrics. These metrics, such as ROUGE \cite{lin2004rouge}, are based on word overlaps with the reference summary. However, word overlap metrics are limited and can weakly correlate with human judgment.

To address these shortcomings, human evaluation is often used, where human annotators assess various aspects of generated summaries. In this tutorial, we will present different kinds of human evaluation experiments, how they are designed, and how they are performed.

\vspace{-1mm}
\subsection{Part V: Future Work [10 min]}
To conclude the tutorial, we will present several notable open questions for opinion summarization, such as the need for additional annotated resources, common issues with the generated summary (e.g., repetition, hallucination, coherency, and factuality),  and the ability to handle various type of input data (e.g., images and knowledge bases). Based on these open questions, we will also present future work on opinion summarization. 

\section{Objectives}
In this tutorial, we will cover a wide range of techniques from pre-neural approaches to the most recent advances for opinion summarization. In addition, we will also introduce the commonly used resources and evaluation metrics. Our goal for this tutorial is to increase the interest of the IR community towards the opinion summarization problem and help researchers to start working on relevant problems. 

\section{Relevance to the IR community}

Sentiment analysis has been a major research area in the IR community. Since the tutorial will cover cutting-edge research in the field, it would attract a wide variety of IR researchers and practitioners. 
We would also like to emphasize that the interest in opinion mining and summarization techniques in the IR community has been rapidly and significantly increased in recent years. To the best of our knowledge, we are the first ones to offer a tutorial covering the series of recent opinion summarization approaches.\footnote{More than 80\% of the papers were published within the last three years.}

\begin{figure}[t]
    \centering
    \includegraphics[width=0.85\linewidth]{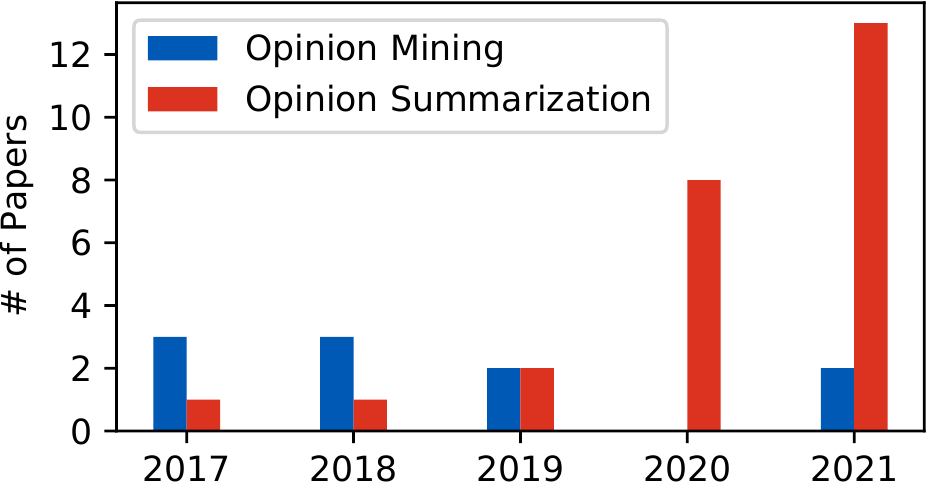}
    \vspace{-0.2cm}
    \caption{Increasing \# of papers for opinion summarization that are published in IR-related venues.}
    \label{fig:stats}
    \vspace{-0.2cm}
\end{figure}

\section{Broader Impact}

Methods presented in the tutorial also have applications beyond customer reviews. The amount of opinions on various topics expressed online is vast. These opinions address various sports, politics, and public events. In turn, this calls for ways to summarize this information for the benefit of the user. As we will discuss, the methods presented in the tutorial can be applied to other opinion domains, such as social media and blogs.

\section{Instructors}
\smallskip

\textit{Reinald Kim Amplayo} is a Research Scientist at Google. He received his PhD from the University of Edinburgh, where his thesis focused on controllable and personalizable opinion summarization. He is a recepient of a best student paper runner-up at ACML 2018.

\smallskip
\textit{Arthur Bra\v{z}inskas} is a Research Scientist at Google working on natural language generation for Google Assistant. His PhD on low- and high-resource opinion summarization is supervised by Ivan Titov and Mirella Lapata at the University of Edinburgh. 

\smallskip
\textit{Yoshi Suhara} is an Applied Research Scientist at Grammarly. Previously, he was a Senior Research Scientist at Megagon Labs, an Adjunct Instructor at New College of Florida, 
a Visiting Scientist at the MIT Media Lab, and a Research Scientist at NTT Laboratories. He received his PhD from Keio University in 2014. His expertise lies in NLP, especially Opinion Mining and Information Extraction.

\smallskip
\textit{Xiaolan Wang} is a Senior Research Scientist at Megagon Labs. 
She received her PhD from University of Massachusetts Amherst in 2019. Her research interests include data integration, data cleaning, and natural language processing. She co-instructed the tutorial,  \textit{Data Augmentation for ML-driven Data Preparation and Integration}, at VLDB 2021. 

\smallskip
\textit{Bing Liu} is a Distinguished Professor of Computer Science at the University of Illinois at Chicago (UIC). He has published extensively in top conferences and journals. He also authored four books about lifelong learning, sentiment analysis and Web mining. Three of his papers received Test-of-Time awards: two from SIGKDD and one from WSDM. He has served as the Chair of ACM SIGKDD from 2013-2017, as program chair of many leading data mining conferences, including KDD, ICDM, CIKM, WSDM, SDM, and PAKDD, and as associate editor of leading journals such as TKDE, TWEB, DMKD and TKDD. He is a recipient of ACM SIGKDD Innovation Award, and he is a Fellow of the ACM, AAAI, and IEEE.

\bibliographystyle{ACM-Reference-Format}
\bibliography{custom,sigir}

\end{document}